\documentclass[letterpaper, 10 pt, conference]{ieeeconf}  

\IEEEoverridecommandlockouts                              

\makeatletter
\let\NAT@parse\undefined
\makeatother

\usepackage{xspace}
\usepackage{amsmath,amssymb,amsfonts,amsfonts}
\usepackage{algorithm}
\usepackage[noend]{algorithmic}
\usepackage{url}
\usepackage{bm}
\usepackage{graphicx}
\usepackage[font=footnotesize]{caption} 
\usepackage[font=footnotesize]{subcaption}
\usepackage{color}
\usepackage{cite}
\usepackage{hyperref}
\usepackage{eqparbox}
\usepackage{multirow}
\usepackage{booktabs}

\title{\LARGE \bf
Generalizing Informed Sampling
for Asymptotically Optimal Sampling-based Kinodynamic Planning via Markov Chain
Monte Carlo
}

\author{
Daqing Yi$^{*1}$,
Rohan Thakker$^{*2}$,
Cole Gulino$^{2}$, 
Oren Salzman$^{2}$ and
Siddhartha Srinivasa$^{1}$
\thanks{$^{*}$Daqing Yi and Rohan Thakker contributed equally to this paper.}
\thanks{$^{1}$Daqing Yi and Siddhartha Srinivasa are with Paul G. Allen School of Computer Science \& Engineering, University of Washington.
{\tt\small \{dqyi, siddh\}@cs.washington.edu}}
\thanks{$^{2}$ Rohan Thakker, Cole Gulino and Oren Salzman are with Robotics Institute, Carnegie Mellon University.
{\tt\small \{rthakker, cgulino, osalzman\} @andrew.cmu.edu}}%
}

\newcommand{\calX}{\ensuremath{\mathcal{X}}\xspace}

\newcommand{\calU}{\ensuremath{\mathcal{U}}\xspace}

\newcommand{\R}{\mathbb{R}}

\newcommand{\Cfree}{\ensuremath{\calX_{\rm free}}\xspace}

\newcommand{\Cinf}{\ensuremath{\calX_{\rm {inf}}}\xspace}

\newcommand{\Pinf}{\ensuremath{\pi_{\hat{f}}}\xspace}

\newcommand{\cbest}{\ensuremath{c_{\rm best}}\xspace}

\newtheorem{prop}{Proposition}

\newcommand{\ignore}[1]{}

\newcommand{\captionstyle}{\sf \footnotesize }

\begin{document}

\maketitle
\thispagestyle{empty}
\pagestyle{empty}

\begin{abstract}
Asymptotically-optimal motion planners such as RRT* have been shown to incrementally approximate the shortest path between start and goal states.
Once an initial solution is found, their performance can be dramatically improved by restricting subsequent samples to regions of the state space that can potentially improve the current solution.
When the motion planning problem lies in a Euclidean space, this region~$\Cinf$, called the informed set, can be sampled directly.
However, when planning with differential constraints in non-Euclidean state spaces, no analytic solutions exists to sampling~$\Cinf$ directly.

State-of-the-art approaches to sampling~$\Cinf$ in such domains such as Hierarchical Rejection Sampling (HRS) may still be slow in high-dimensional state space.
This may cause the planning algorithm to spend most of its time trying to produces samples in~$\Cinf$ rather than explore it.
In this paper, we suggest an alternative approach to produce samples in the informed set~\Cinf for a wide range of settings.
Our main insight is to recast this problem as one of sampling uniformly within the sub-level-set of an implicit non-convex function.
This recasting enables us to apply Monte Carlo sampling methods, used very effectively in the Machine Learning and Optimization communities, to solve our problem.
We show for a wide range of scenarios that using our sampler can accelerate the convergence rate to high-quality solutions in high-dimensional problems.
\end{abstract}

\section{Introduction}
\label{sec:intro}

Sampling-based motion-planning algorithms~\cite{L06} have proven to be an effective tool at solving motion-planning problems.
They search through a continuous state space~$\calX$ by sampling random states and maintaining a discrete graph~$G$ called a \emph{roadmap}.
Vertices and edges in $G$ correspond to collision-free states and paths, respectively.

Roughly speaking, these algorithms iteratively sample new states.
This is required to ensure that, as the number of samples tends to infinity, 
(i)~a solution will be found 
and that
(ii)~given some optimization criteria, the quality of the solution will progressively converge to the quality of the optimal solution.

Initially, 
when a path has yet to be found, 
the samples are drawn from the entire state space~$\calX$.
However, once a path $\gamma$ is produced,  algorithms that seek \emph{high-quality paths} can limit their sampling domain to a subset of~$\calX$ only  containing states that may be used to produce higher-quality paths than~$\gamma$.
Following Gammell et al.~\cite{GSB14}, we call this subset the \emph{informed subset} and denote it~$\Cinf$.
In this work we address the problem of efficiently producing samples in informed subset for systems with arbitrary complex costs. 

For Euclidean spaces optimizing for path length, 
\Cinf can be analytically expressed as a prolate hyperspheroid and can be sampled directly using a closed-form solution~\cite{GSB14}.
Indeed, directly sampling in \Cinf has been shown to dramatically improve computation time when compared to sampling in~$\calX$, especially in high dimensions. 

Unfortunately, in more general settings,
it is not clear how to directly sample \Cinf.
One approach to produce samples in~\Cinf is via \emph{rejection sampling}---sampling a state $x \in \calX$ and testing if $x \in \Cinf$.
However, when the size of the informed space~$\Cinf$ is much smaller than entire state space~$\calX$, this procedure is highly inefficient, dominating the running time of the algorithm~\cite{KTC16}.
Recently, Kunz et al.~\cite{KTC16} showed, under some technical assumptions, how to partially ameliorate this inefficiently by \emph{Hierarchical rejection sampling} (HRS). 
Here, individual dimensions are sampled recursively 
and then combined. Rejection sampling is performed for these partial samples until a suitable sample has been produced. 
Unfortunately, HRS may still produce a large number of rejected samples especially in high-dimensional spaces~\cite{KTC16}.
This may cause the planning algorithm to spend most of its time trying to produces samples in~$\Cinf$ rather than explore it.

\begin{figure}[tb]
  \centering
  	\includegraphics[width=0.75\linewidth ]{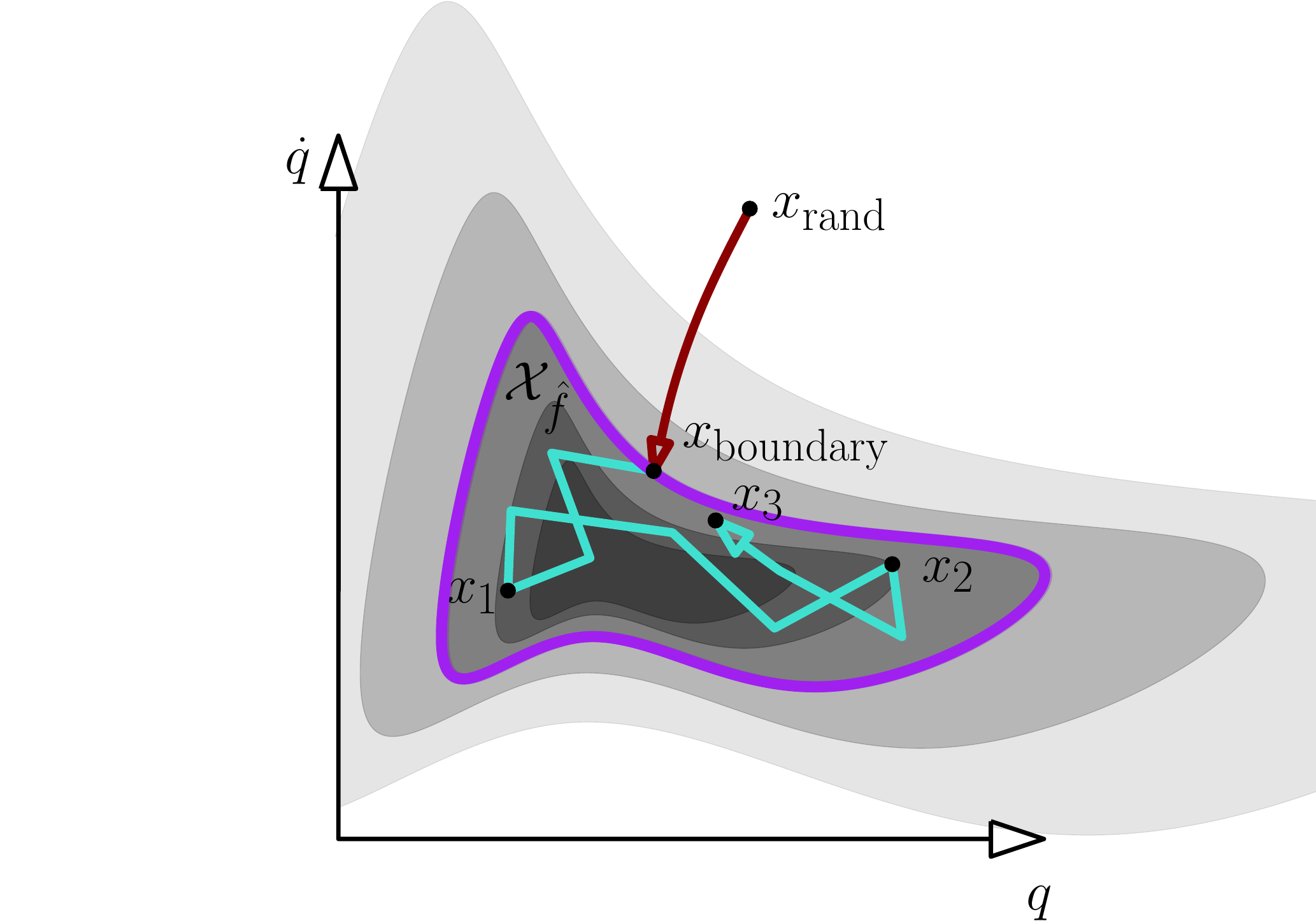}
  \caption{
    \captionstyle
  	Algorithmic approach.
  	Cost function is depicted using isocontours (darker shades reflect lower cost) while the boundary of the informed set is depicted in purple. 
  	The root-finding and MCMC algorithms are depicted in red and turquoise, respectively.
  	}
   	\label{fig:alg}
\end{figure}

In this paper, we suggest an alternative approach to produce samples in the informed set \Cinf for a wide range of settings.
\textbf{
Our main insight is to recast this problem as one of sampling uniformly within the sub-level-set of an implicit non-convex function.
This recasting enables us to apply Monte Carlo sampling methods, used very effectively in the Machine Learning and Optimization communities, to solve our problem.
}
Specifically, our approach, depicted in Fig~\ref{fig:alg} consists of two stages:
in the first, a random sample $x \in \calX$ is retracted to the boundary of~$\Cinf$ by running a root-finding algorithm;
in the second stage, this retracted sample  is used to seed a Monte Carlo sampling chain which allows us to  produce samples that (approximately) cover~$\Cinf$  uniformly.

While our approach can be used with any Markov Chain Monte Carlo (MCMC) method, it is especially suited to be used with Hit-and-Run~\cite{KSZ11}.
Roughly speaking, this is because Hit-and-Run (detailed in Sec~\ref{sec:algorithm})
produces a series of one-dimensional rejection samples which are extremely fast to compute, even in high-dimensional spaces. 

Our approach requires that the system has a solution to the two-point boundary value problem (2pBVP)~\cite{L06} and that a gradient can be defined over the cost function.
Indeed, we demonstrate the efficiency of our approach on a wide variety of systems and show that it enables reducing the planning time by several orders of magnitude when compared to algorithms using rejection sampling or HRS.

\begin{figure*}[t!]
	\centering
	\begin{minipage}[t]{0.64\linewidth}
	\begin{subfigure}[b]{\textwidth}
		\centering
		\includegraphics[height=2.7cm]{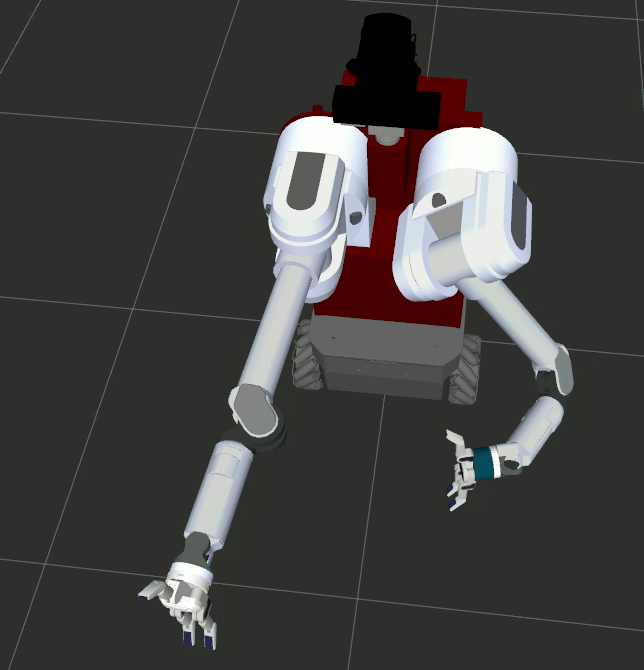}
		\includegraphics[height=2.7cm]{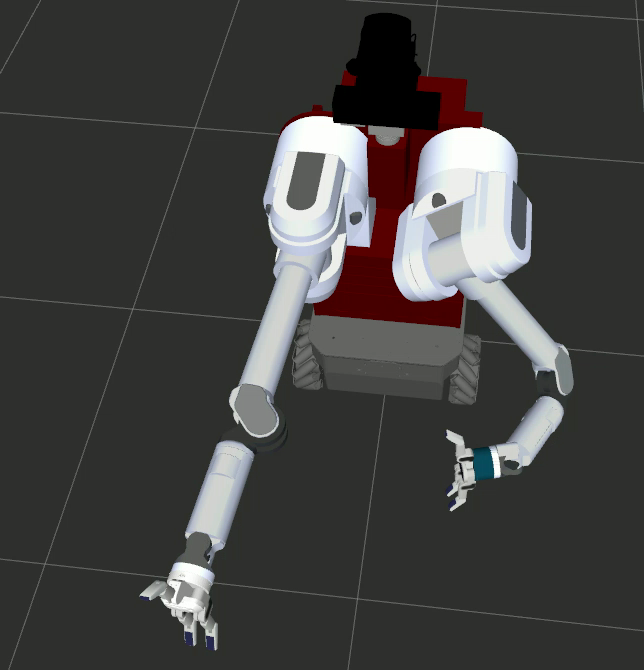}
		\includegraphics[height=2.7cm]{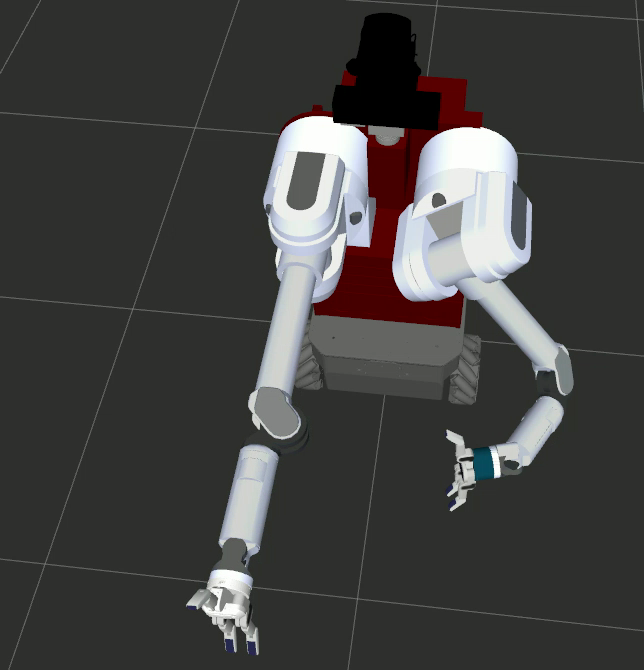}
		\includegraphics[height=2.7cm]{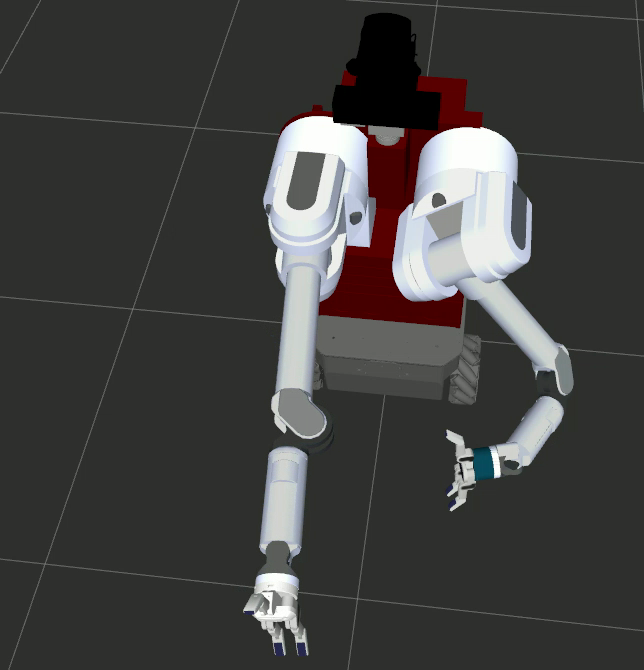}
		\caption{Both start velocity and goal velocity are zero.}	\vspace{4pt}
		\label{fig:motivation:slow}
	\end{subfigure}
	\begin{subfigure}[b]{\textwidth}
		\centering
		\includegraphics[height=2.7cm]{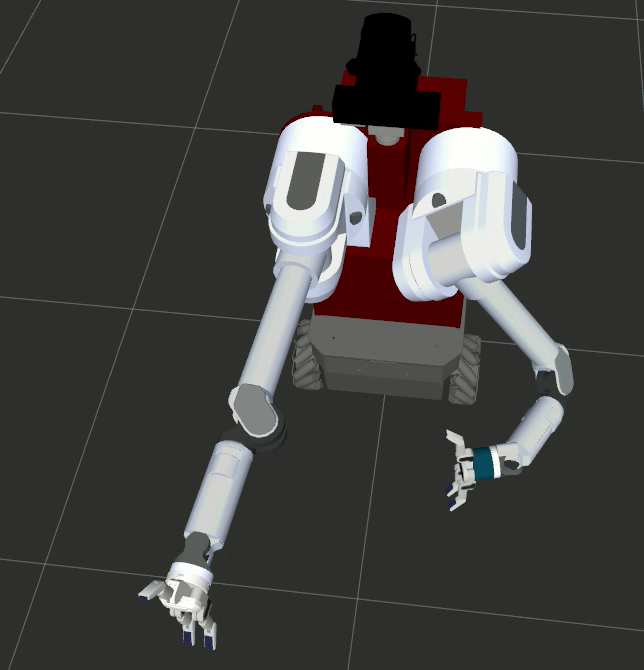}
		\includegraphics[height=2.7cm]{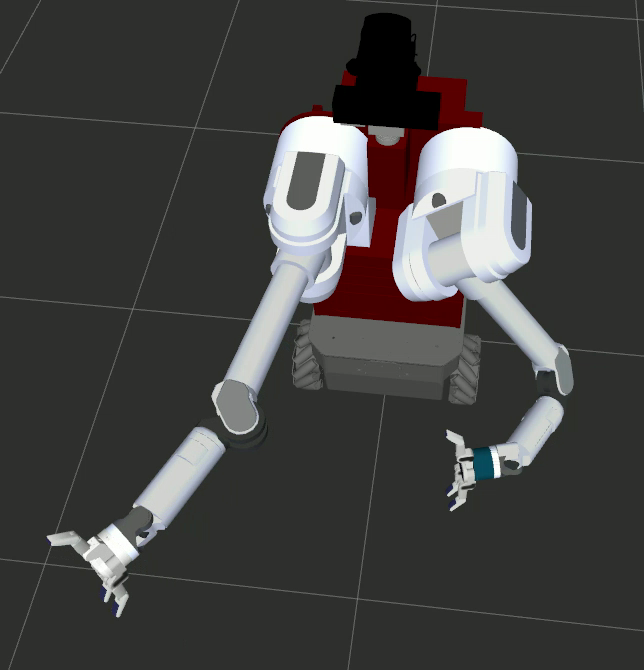}
		\includegraphics[height=2.7cm]{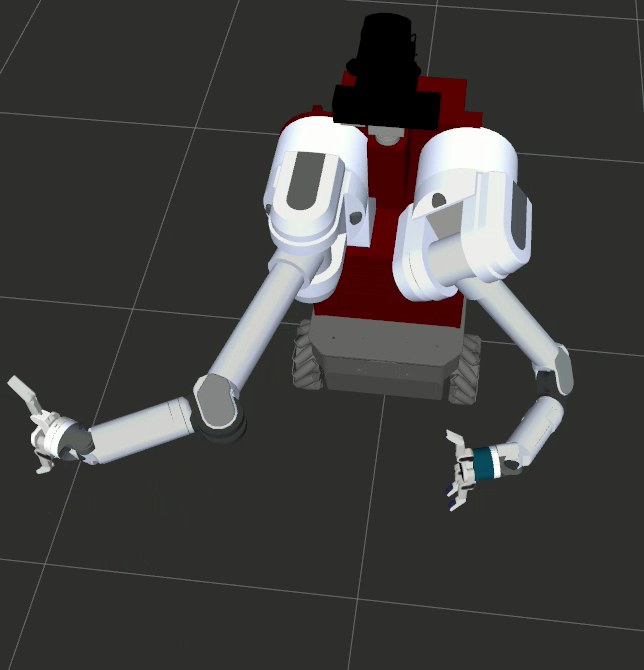}
		\includegraphics[height=2.7cm]{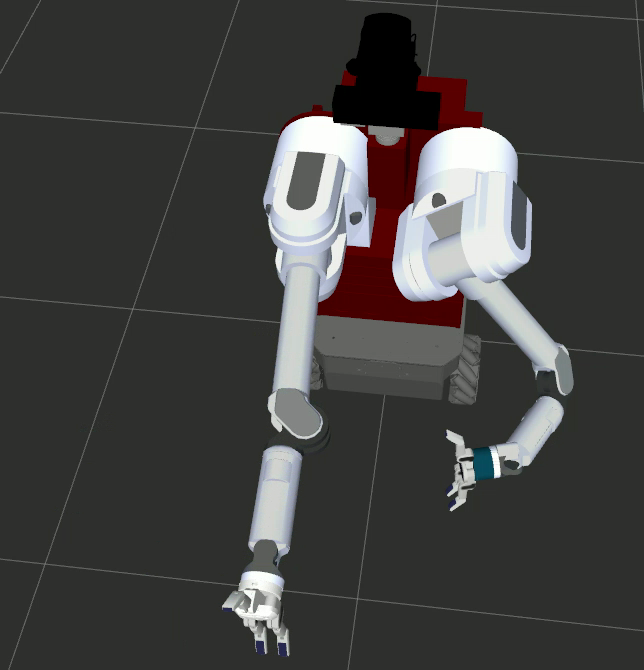}
		\caption{Start velocity is zero but goal velocity is non-zero.}
		\label{fig:motivation:fast}
	\end{subfigure}
    \end{minipage}
    \begin{minipage}[t]{0.35\linewidth}
    	\begin{subfigure}[c]{\textwidth}
    		\includegraphics[width=\textwidth]{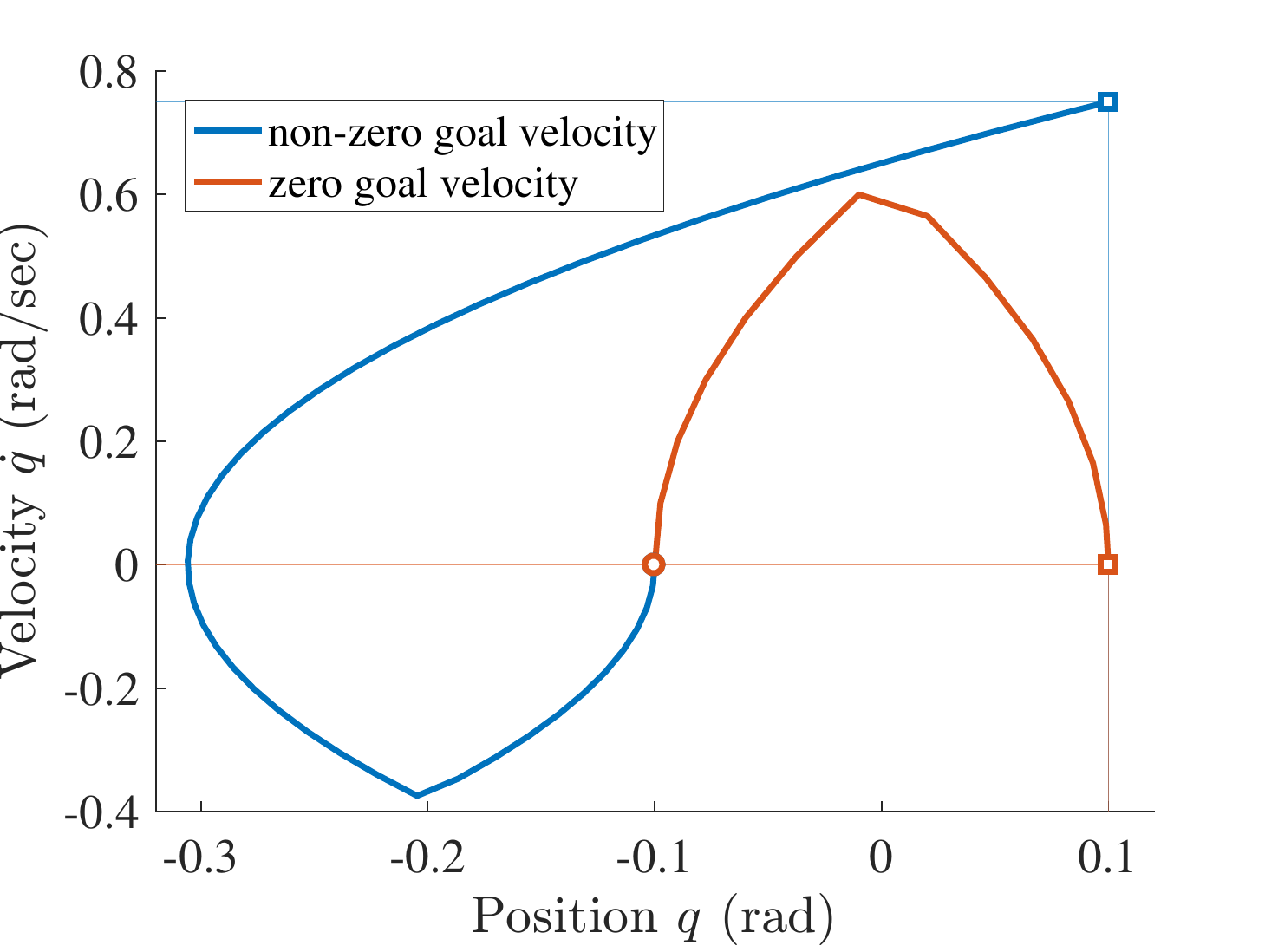}
    		\caption{Phase plot of two trajectories of one of the joints.}
    		\label{fig:motivation:phase_plot}
        \end{subfigure}
   	\end{minipage}
	\caption{HERB moves right arm from a start configuration to a goal configuration, which are in close proximity.
    When the goal velocity is non-zero, HERB needs to move right arm further away to accelerate.
	}
	\label{fig:motivation}
\end{figure*} 

The rest of the paper is structured as follows: after describing related work in Sec.~\ref{sec:related_work}, we formally define our problem in Sec.~\ref{sec:pdef}.
We then provide in Sec.~\ref{sec:mtdi} an intuitive description of the challenges faced in sampling within the informed set for our planning domains.
We continue in Sec.~\ref{sec:algorithm} with a description of our algorithm and present experimental evaluations in Sec.~\ref{sec:eval}.
Finally, we conclude with a discussion in Sec.~\ref{sec:future}.

\section{Related work}
\label{sec:related_work}
We start in Sec.~\ref{subsec:planning} by giving an overview of relevant sampling-based motion-planning algorithms.
We then continue in Sec.~\ref{subsec:sampling} to describe different approaches that can be used by  these algorithms to sample~$\calX$.
We conclude our literature review in Sec.~\ref{subsec:mcmc} with a brief overview of Markov Chain Monte Carlo methods.

\subsection{Sampling-based motion-planning algorithms}
\label{subsec:planning}
Initial sampling-based algorithms such as RRT~\cite{LK01} and PRM~\cite{KSLO96} did not take into account the \emph{quality} of a path, given some optimization criteria, and only guaranteed to asymptotically return \emph{a} solution, if one exists.
Karaman and Frazzoli~\cite{KF11}, presented variants of PRM and RRT, named PRM* and RRT*, respectively that were shown to produce paths who's cost converges asymptotically to the minimal-cost path.
This was done by recognizing the underlying connections between stochastic sampling-based motion planning and the theory of random geometric graphs (see also~\cite{SSH16}).
Additional algorithms followed, increasing the converges rate by various techniques such as 
lazy dynamic programming~\cite{GSB15,SH15},
relaxing optimality to near-optimality~\cite{DB14,SH16} 
and more.

Many of the algorithms mentioned require solving a two-point boundary value problem (2pBVP) to perform exact and optimal connections between vertices in the roadmap.
For holonomic robots, these are simply straight lines in the configuration space, but for kinodynamic sytems with arbitrary cost functions,  computing an optimal trajectory between two states is non-trivial in general.

Xie et al.~\cite{XBPA15} use a variant of sequential quadratic programming (SQP) to solve 2pBVP and integrate it with BIT*~\cite{GSB15}.
Webb and van den Berg~\cite{WB13} use a fixed-final-state-free-final-time controller to solve the 2pBVP  with respect to a cost function that allows for balancing between the duration of the trajectory and the expended control effort.
Perez et al.~\cite{PPKKL12} propose a variant of RRT* that automatically defines a distance metric and node extension method by locally linearizing
the domain dynamics and applying linear quadratic regulation (LQR).

Finally, we note that we are not the first to integrate Monte Carlo sampling into planning algorithms. 
T-RRT~\cite{JCS10} and its variants~\cite{DSC13} are inspired by Monte Carlo optimization techniques and use notions such as the Metropolis criterion~\cite{CG95} to guide the exploration of the configuration space.

\subsection{State-space sampling}
\label{subsec:sampling}
There is a rich body of literature on how to produce samples that increase the efficiency of a planner in terms of finding a solution or producing high-quality solutions.
Heuristic approaches include
sampling on the medial axis~\cite{WAS99a, YDLTA14},
sampling near the boundary of the obstacles~\cite{YTEA12},
resampling along a given trajectory~\cite{AS11}
and more~\cite{US03, SWT09}.
For planning under the differential constraints,
reachability-guided sampling~\cite{PLAEFRA17} focuses on sampling regions of the state space that are most likely to promote expansion for the given constraints.

Of specific interest to our work are approaches that produce samples in the informed set~\Cinf.
As mentioned in Sec.~\ref{sec:intro} Gammel et al.~\cite{GSB14} describe an approach to sample uniformly in~\Cinf for the specific case where $\calX = \R^d$ and when optimizing for path length.
To the best of our knowledge, the only method to produce samples in non-Euclidean spaces that can be  applied to motion planning problems (other than rejection sampling) is HRS by Kunz et al.~\cite{KTC16}.

\subsection{Markov Chain Monte Carlo (MCMC)}
\label{subsec:mcmc}
Monte Carlo simulation is a general sampling framework widely used in various domains. 
Roughly speaking, Monte Carlo simulation repeatedly samples a domain at random to approximate some value or function.
One specific domain where Monte Carlo simulation is used which is relevant to this work is generating draws from a desired distribution which is hard to sample directly.

One of the popular classes of Monte Carlo simulation is 
\emph{Markov Chain Monte Carlo} (MCMC)~\cite{ADDJ03}.
Here, the samples are drawn by generating a Markov chain such that the distribution of points on the chain converges to the desired distribution.
One variant, which is of special interest to us is 
Hit-and-Run~\cite{KSZ11}.
Here, given the current point $x_i$ the next point~$x_{i+1}$ in the Markov chain is produced by sampling a random direction~$\theta$ on the surface of the unit sphere centered at $x_{i+1}$. This defines a ray $r_i$ rooted at $x_i$ and passing through~$\theta$. The point $x_{i+1}$ is chosen by randomly sampling a point on $r_i$.
This algorithm is considered to be one of the most efficient algorithms for generating an asymptotically uniform point if the set under consideration is convex~\cite{LV06}
and it can also be extended to sample points that converge to an arbitrary target distribution in total variation~\cite{RS94}.

The attractiveness of Hit-and-Run for our problem domain stems from the fact that it performs a series of one-dimensional rejection samples which are extremely fast to compute, even in high-dimensional spaces. 
Finally, it is worth noting that we are not the first to apply Hit-and-Run for motion-planning problems.
Recently~\cite{YPVA17} was used as an alternative to RRT to produce \emph{feasible motions} (and not high-quality paths). 
Interestingly the paper concludes with the statement ``\emph{One drawback is that the sample paths for Hit-and-Run have no pruning and are therefore longer than the RRT paths. Hybrid approaches that yield short paths but also explore quickly are a promising future direction.}''
Our paper can be seen as a hybrid approach marrying sampling-based planning with MCMC-based approaches.

\section{Problem definition}
\label{sec:pdef}

Let $\calX, \calU$ denote the state and controls spaces, respectively and set $\Cfree \subset \calX$ to be the set of states where the robot is collision free.
A \emph{trajectory} $\gamma$ is a timed path through~$\calX$ obtained by applying at time $t$ control $u(t) \in \calU$ and satisfying the system dynamics 
$\dot{x}(t) = f( x(t) , u(t) )$.
A trajectory is collision free if $\forall t,~\gamma(t) \in \Cfree$

Given a cost function $C : \calX \times \calU \rightarrow \R$, the cost of a trajectory $ \gamma $ is the accumulated cost along the path
$c(\gamma) = \int_0^{T} c( x(t), u(t) ) |\dot{\gamma}(t)|dt$, 
where $T$ is the duration of~$\gamma$.

Given start and target states $x_s, x_g \in \calX$, we wish to find a collision free trajectory $\gamma^*$ connecting $x_s$ to $x_g$ such that 
$c(\gamma^*) = \min_{\gamma \in \Gamma} c(\gamma)$, where $\Gamma$ is the set of all collision-free trajectories.

Given a trajectory $\gamma_{\rm best}$ with cost $\cbest = c(\gamma_{\rm best})$ the \emph{informed set}~\Cinf is defined to be all states $x$  which may be on trajectories with lower cost than $\cbest$.
Specifically,
$
\Cinf = \{ x \in \calX \mid  
		c ( \gamma^*(x) ) < \cbest \} $~\cite{GSB14}.
Here~$ \gamma^*(x) $ denotes the optimal trajectory  from $ x_s $ to $ x_g $ constrained to pass through $ x $.
Notice that we do not require that~$ \gamma^*(x) $ is collision free.

In this work we consider the problem of efficiently producing samples within~$\Cinf$.
These samples will be used within the informed RRT* framework to efficiently and incrementally compute trajectories of decreasing cost, converging to the optimal trajectory.

\section{Motivation---\Cinf in kinodynamic state spaces}
\label{sec:mtdi}

In this section we properly motivate this work.
Specifically, we start by describing the differences in between planning in Euclidean configuration spaces (also called geometric planning) nd non-Euclidean  state spaces.

\subsection{Geometric vs. Kinodynamic planning}
Consider the problem depicted in Fig.~\ref{fig:motivation} where HERB is required to produce a large velocity at the end of its arm at the goal position.
One approach to address this problem is to first plan in the geometric configuration space and then re-scale the trajectory in time.
However, when the start and goal are in close proximity, a geometric planner will simply connect the two states (Fig.~\ref{fig:motivation:fast}).
On re-scaling this trajectory in time, reaching the goal velocity in such short distance will require large acceleration, which will not be feasible.
Hence, it is required to move the arm back and then reach the goal, i.e. the trajectory returned by the kinodynamic planner shown in (Fig.~\ref{fig:motivation:fast}).
The difference between the two motions are shown in a phase plot in Fig.~\ref{fig:motivation:phase_plot}.


\subsection{Minimal Time Double Integrator}
To understand why we resort to optimization-based methods and do not attempt to provide a closed-form solution to sample $\Cinf$ we study the structure of the informed set for a simple yet important dynamical system---the double integrator minimizing time (MTDI). 
Here, we are given a one-dimensional point robot with bounded acceleration moving amid obstacles. We wish to compute the minimal-time trajectory between two states $x_s, x_g$.
A state $x \in \calX$ in this model is defined by 
the position $q \in \R$
and
the velocity $\dot{q}\in \R$ of the robot.
The system dynamics are described by:
\begin{equation}
\begin{bmatrix}
	\dot{q} \\
	\ddot{q}
\end{bmatrix}
=
\begin{bmatrix}
	0 & 1 \\
	0 & 0
\end{bmatrix}
\begin{bmatrix}
	{q} \\
	\dot{q}
\end{bmatrix}
+
\begin{bmatrix}
	0 \\
	1
\end{bmatrix}
u.
\end{equation}
Here, the control 
$u \in [\underline{u}, \overline{u}]$ 
is the (bounded) acceleration.

Notice that 
(i)~this is model can be seen as a simplified one-dimensional instance of a robot manipulator with many degrees of freedom and that
(ii)~closed-form solutions exist to the 2pBVP for this specific case (as well as the multi-dimensional setting)~\cite{HN10, KS14}.

Recall that for Euclidean spaces minimizing path length, the informed set~\Cinf is a prolate hyperspheroid~\cite{GSB14}.
Moreover, the size and shape of the hyperspheroid is defined only be the cost $\cbest$ of the current best solution and not by the location of the start~$x_s$ and goal~$x_g$.

For the case of a MTDI, this is not the case. 
Specifically, we have that 
(i)~the structure of~\Cinf changes not only with~$\cbest$ but also according to the specific values of~$x_s$ and~$x_g$ 
and that
(ii)~the cost map that implicitly defines~\Cinf can contain discontinuities (in contrast to Euclidean spaces minimizing path length where the cost map is continuous and differentiable at every point).

\begin{figure}[tb]
  \centering
  	\includegraphics[height = 5.25cm ]{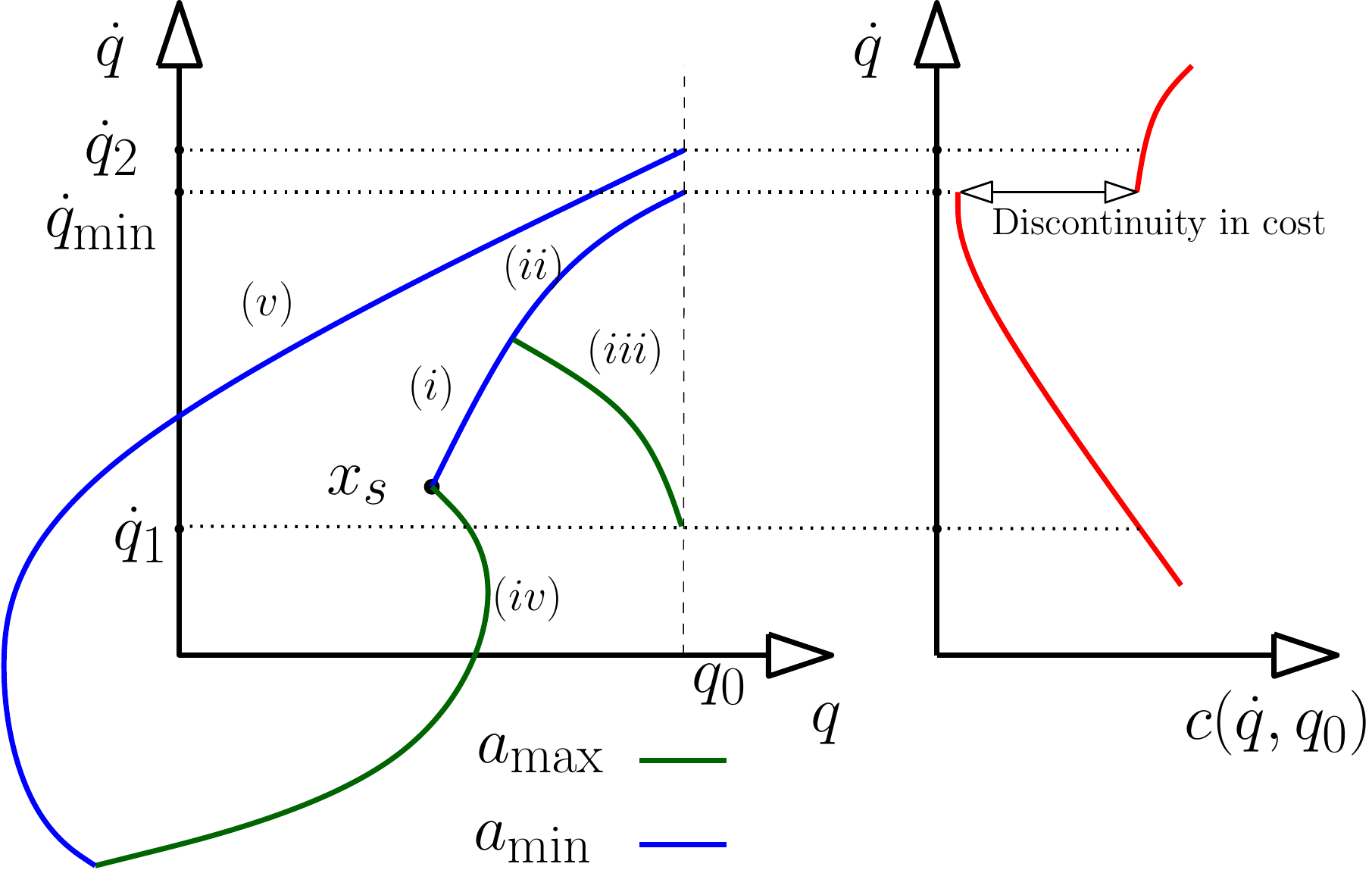}
  \caption{
    \captionstyle
  	Visualization of the discontinuity in the cost function of MTDI (right) related to the types of controls applied (left). 
  	Given state~$x_s$ and fixed position $q_0$, we depict the cost (time) as a function of the velocity~$\dot{q}$. 
  	The minimal cost is attained at $\dot{q}_{\min}$ by applying maximal acceleration (blue curves~$(i), (ii)$). 
  	To reach states such as~$\dot{q}_1$, where $\dot{q}_1 < \dot{q}_{\min}$ we need to apply maximal acceleration (curve~$(i)$) followed by minimal acceleration (green curve~$(iii)$), which result in a continuous increase in cost.
  	However, for states such as $\dot{q}_2$, where $\dot{q}_2 > \dot{q}_{\min}$, we need to apply minimal acceleration  followed by maximal acceleration (curves~$(iv), (v)$), which result in the discontinuity.
  	}
   	\label{fig:discont}
\end{figure}

To understand the differences recall that optimal trajectories  for MTDI follow a ``bang-bang'' controller~\cite{HN10, KS14}.
Namely, we first apply maximal (or minimal) acceleration for some duration and then switch to applying minimal (or maximal, respectively) acceleration.
It is straightforward to see that both the type and the amount of acceleration applied (and hence the structure of~\Cinf) depend on the specific values of~$x_s$ and~$x_g$. 
Fig~\ref{fig:discont} depicts a simple example where the cost map is discontinuous.

To summarize, the structure of \Cinf can change given different start and goal states.
Furthermore,  its boundary may not be  differentiable due to the aforementioned discontinuous.

\section{MCMC-based Informed Sampling}
\label{sec:algorithm}

In this section we describe our approach to efficiently produce new samples in an informed set~\Cinf given a specific cost~$\cbest$ of trajectory $\gamma_{\rm best}(t)$.
The samples follow a Markov Chain Monte Carlo, in which a new sample candidate is produced from a previous sample that also lies in the same informed set.
Furthermore, the value~$\cbest$ can decrease between consecutive iterations in the planning process of an informed RRT* planner.
This will occur if the search algorithm that uses the sampler finds a path to the goal whose cost is lower than~$\cbest$.

The idea behind applying MCMC for informed sampling is to define a target distribution~$ \pi $ that has $p(x_{sample}\in\Cinf) \ne 0 ~\&~ p(x_{sample})\notin\Cinf = 0$. 
This is specially useful if we want to bias the samples based on our knowledge of the environment.
However, we make no such assumption about the environment and use a uniform distribution over all points in~$ \Cinf $. 
Our approach consists of two stages,
\begin{enumerate}
	\item finding an initial sample $ x_0 \in \Cinf $ which will serve as the start of a Markov chain. 
	This is implemented using the function 
	$ \texttt{sample\_in\_informed\_space}( )$, and
	\item sampling a new sample $ x_i \in \Cinf $ given a previous sample $ x_{i-1} $. 
	This is implemented using the function $\texttt{MCMC\_sample} (x_{i-1}, \cbest)$.
\end{enumerate}
Our framework is described in Algorithm~\ref{alg:mcmc_informed_sampling} and visualized in Fig.~\ref{fig:alg}.
We now continue to detail each of the algorithm's stages.

\subsection{Finding an initial sample in $\Cinf$}
In theory, MCMC methods converge to the desired distribution regardless of the initial sample used to seed the chain.
In our setting, the probability distribution~\Pinf is defined by having all points in \Cinf distributed uniformly
while 
the probability of sampling any configuration $x \in \calX \setminus \Cinf$ is zero.
A common practice to avoid starting biases in MCMC-type algorithm is to discard an initial set of samples (a process referred to as ``burn-in'')~\cite{ADDJ03}. 

In our setting, we are only interested in points in~\Cinf, thus we suggest to start the Markov Chain in~\Cinf and avoid this burn-in stage. 
We restart our process and generate a new Markov chain  when 
(i)~the cost of $\cbest$ is updated (i.e. a new solution is found by the planner) or
(ii)~the new sample on the existing Markov chain is outside~\Cinf.

We suggest several methods to produce an initial sample~$x_0 \in \Cinf $ 
\begin{itemize}
	\item randomly returning either the start state or the goal state,
	\item randomly sampling a state $x_{\text rand} \in \Cinf$ and using a gradient descent algorithm (e.g. Newton-Raphson Method~\cite{RT06}) to find a sample in~$ \Cinf $
	\item sampling from a pool of previous samples that are in the informed set $ \Cinf $ and
	\item applying rejection sampling until a sample in the informed set is found.
\end{itemize}

Each of the methods proposed has its own pros and cons.
For example, a gradient-descent algorithm is usually efficient in finding a solution, but subject to only convex problems.
Sampling from a pool of samples is algorithmic-free but biases new samples to be near previous samples.


\subsection{Generating a new sample in a Markov chain}
\label{mcmc}

Our approach is general and can be applied to any MCMC algorithm (see Sec.~\ref{sec:related_work}).
The process is demonstrated in Algorithm~\ref{alg:mcmc_informed_sampling}.
At the beginning of a Markov chain, \texttt{sample\_in\_informed\_space}() is called to generate the first sample in an informed set.
\texttt{MCMC\_sample}() is called to generate a new sample based on a previous sample $ x_{i-1} $ and a cost $ \cbest $ that defines an informed set.
We demonstrate how to instantiate it with two different algorithms \emph{Metropolis-Hastings} and \emph{Hit-and-Run}, which will be described in later subsections.
If a generated new sample candidate is in the informed set, this candidate will be returned as a new sample.
But if a generated new sample candidate is not in the informed set, it will go back to line 2.
A new Markov chain will be initiated by calling \texttt{sample\_in\_informed\_space}() to generate a new sample $ x_0 $.

\begin{algorithm}[t]
	\begin{algorithmic}[1]
		\LOOP 
		\IF{ $ x_{i-1} == \varnothing $ }  
		\label{alg:mcmc_informed_sampling:start}
			\STATE $x_{0} \leftarrow \texttt{sample\_in\_informed\_space}( )$
		\ENDIF
		\STATE $x_{i} \leftarrow \texttt{MCMC\_sample} (x_{i-1}, \cbest)$ 
		\IF{ $ x_{i} \not\in \Cinf $ }
			\STATE $ i \leftarrow 0 $; $ x_{0} \leftarrow \varnothing $
			\STATE \textbf{Goto} line \ref{alg:mcmc_informed_sampling:start} 
		\ENDIF
		\STATE \textbf{return} $x_{i}$
		\ENDLOOP 
	\end{algorithmic}
	\caption{\captionstyle MCMC-based Informed Sampling $( x_{i-1}, \cbest)$}
	\label{alg:mcmc_informed_sampling}	
\end{algorithm}

\subsubsection{Metropolis-Hastings sampler}


The Metropolis-Hastings algorithm is one of the most popular MCMC samplers~\cite{CG95}, because it provides a simple and parameter-free framework that guarantees the convergence of Markov chains to a target distribution.
Our work adopts the general Metropolis-Hastings algorithm, as described in Algorithm~\ref{alg:mh_mcmc}, 
We generate a new sample $ x_{i}$ around the previous sample $ x_{i-1}$ using a Gaussian distribution (line~1).
We then check if the point lies in the informed set (lines 2-5) and if it does we return it.
If not, we return the previous sample.
An acceptance ratio $ \alpha $ is used to keep the reversibility even if the target probability $ \pi $ is asymmetric, which is needed to guarantee the convergence~\cite{CG95}. 


\begin{algorithm}[t]
	\begin{algorithmic}[1]
		\STATE $ x'_{i} \leftarrow \texttt{sample\_normal}( q ( x \mid x_{i-1},\Sigma) ) $ 
		\label{start}
		\STATE $ \alpha \leftarrow \frac{ q ( x_{i-1} \mid x'_{i},\Sigma) \pi( x'_{i} ) }{ q ( x'_{i} \mid x_{i-1},\Sigma) \pi( x_{i-1} ) } $
		\IF{ \texttt{sample\_random}$ (0.0 , 1.0) > \min (1, \alpha) $  }
            \RETURN $ x'_{i} $
		\ENDIF
		\RETURN $ x_{i} $
	\end{algorithmic}
	\caption{\captionstyle Metropolis-Hastings MCMC $(x_{i-1}, \cbest)$}
	\label{alg:mh_mcmc}	
\end{algorithm}

In implementation, we use Newton-Raphson method as a gradient descent with random restart to find $ x_0 \in \Cinf $ as the start of a Markov chain.

\subsubsection{Hit-and-Run sampler}

Hit-and-Run~\cite{S84} sampler is known to efficiently generate uniform samples. 
Specifically, we use the Accelerated Hit-and-Run variant~\cite{KSZ11} of the algorithm i.e. described in Algorithm~\ref{alg:hit_and_run_mcmc}, which also supports the uniform sampling in both convex and non-convex state space~\cite{KSZ11}.
Given the previous sample~$x_{i-1}$ it first  samples a random direction on a unit sphere (line 1).
This induces a line~$L(\lambda)$ passing through~$x_{i-1}$ in the direction sampled (line 2), parametrized by a scalar~$\lambda$.
We obtain upper and lower bounds on $\lambda$ (line 3) that are problem dependent. 
For example, if we have box constraints on joint limits of the robot and on maximum velocity, then bounds are given by $\lambda^{+} = -\lambda^{-} = l_{diag}$; where $l_{diag}$ is the length of the longest diagonal of the box.
We then sample a point along~$L(\lambda)$ by sampling a scalar~$\lambda'$ within our bounds (line 5).
This defines a point~$x_{i}$ which is a candidate for the next sample along the Markov Chain (line 6).
We then check if the point lies in the informed set (line 7) and if it does, we return it.
If not, we update our bounds (lines~9-12) and repeat the process.
The algorithm can be viewed as an efficient method that performs rejection sampling along a one-dimensional line passing through the previous sample parametrized by $\lambda$.

\begin{algorithm}[t]
	\begin{algorithmic}[1]
		\STATE $d \leftarrow$ \texttt{sample\_random\_direction}$()$
		\STATE $ L(\lambda) = \{  x \mid x = x_{i-1} + \lambda d_i \} $
		\STATE $ \lambda^{+} \leftarrow \sup L(\lambda) $; 
		\hspace{3mm} 
		$ \lambda^{-} \leftarrow \inf L(\lambda) $
		
		\LOOP
		
		\STATE $ \lambda' \leftarrow$ \texttt{sample\_random}$ (\lambda^{-} , \lambda^{+})$
		\STATE $ x_{i} \leftarrow x_{i-1} + \lambda'_{i} d_i $
		
		\IF{ $ c( x_{i} ) $ $ < \cbest $ }
		\RETURN $ x_{i}$
		\ENDIF
		
		\IF{ $ \lambda' > 0 $ }
		\STATE $ \lambda^{+} \leftarrow \lambda' $
		\ELSE
		\STATE $ \lambda^{-} \leftarrow \lambda'$
		\ENDIF
		
		\ENDLOOP
	\end{algorithmic}
	\caption{\captionstyle Hit-and-Run MCMC $(x_{i-1}, \cbest)$}
	\label{alg:hit_and_run_mcmc}	
\end{algorithm}

For this algorithm, we continue sampling along a Markov Chain until either 
(i)~the difference between the lower and upper bounds ($\lambda^-$ and $\lambda^+)$ 
that define our sampling domain is below a predefined threshold or
(ii)~a predefined number of samples was exceeded.
We want to point out that a Hit-and-Run sampler only requires that a Markov chain starts in an informed set, and will not produce a sample outside of the informed set.
Also, in our implementation, we pick the start or the goal state to find $ x_0 \in \Cinf $ as the start of a Markov chain.

\subsection{Asymptotic optimality}
We note that our approach produces samples that cover the informed space. 
Namely, there is a non-zero probability to sample in any region of $\Cinf$.
A direct implication of the proof of optimality presented in~\cite{KF11} is that our algorithm is asymptotic optimal:

\begin{prop}
	\label{prop:asym_opt}
	Informed RRT*~\cite{GSB14} running with MCMC-based informed sampling is asymptotic optimal.	
\end{prop}

\section{Evaluation}
\label{sec:eval}

We evaluate the performance of proposed MCMC methods by comparing four types of samplers, which are Rejection Sampler (RS), Hierarchical Rejection Sampler (HRS), Metropolis-Hastings Sampler (MH), and Hit-and-Run Sampler (HNR).
We use different samplers to generate a fixed number of samples in different informed sets to check the sampling efficiency. 
We then compare the quality of the samplers by evaluating how the samplers work with informed RRT*~\cite{GSB14}.

\subsection{Sampling Efficiency}

\begin{figure*}[t!]
	\centering
	\begin{subfigure}[t!]{0.325\textwidth}
		\centering
		\includegraphics[width=\textwidth]{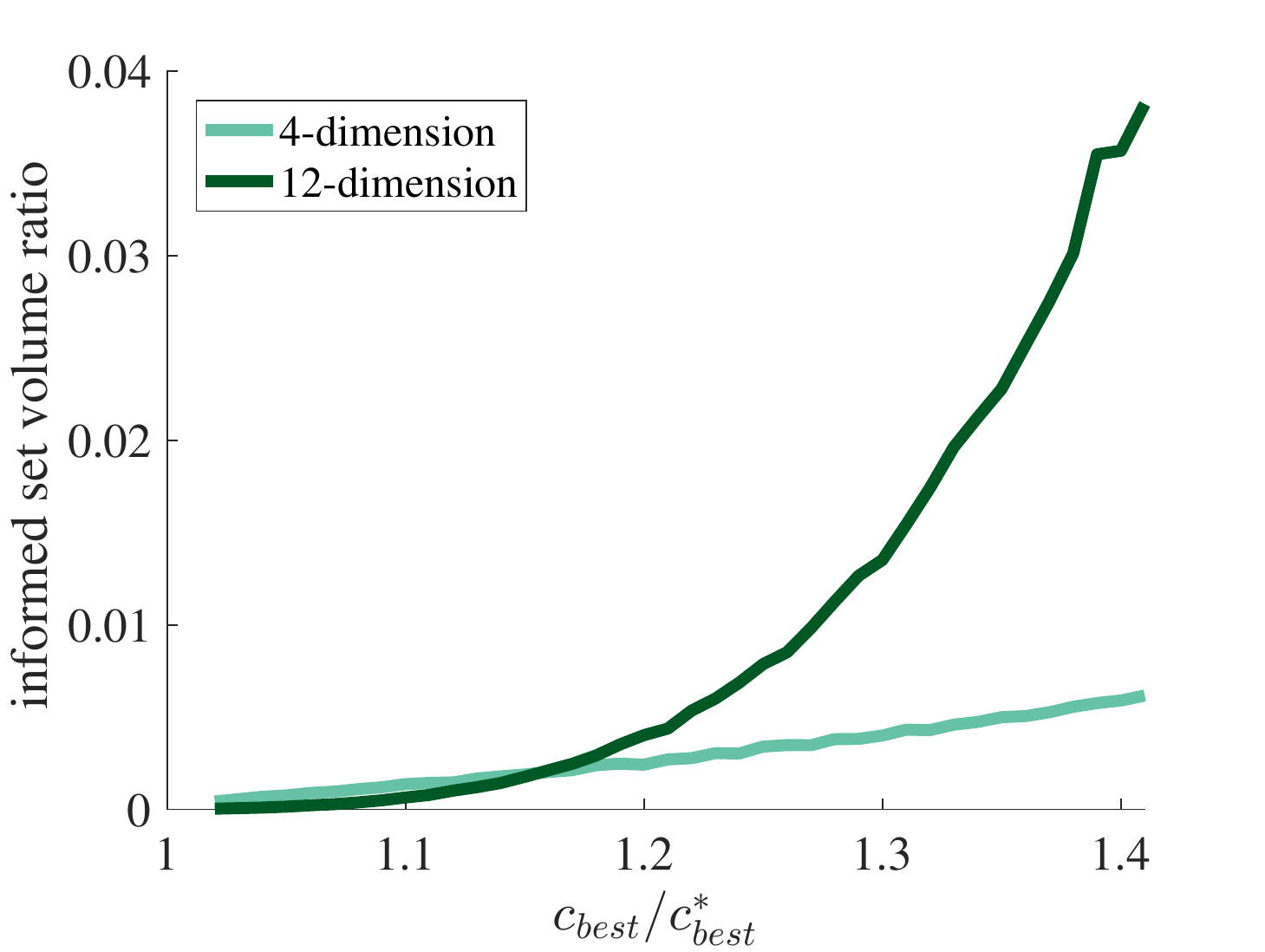}
		\caption{\captionstyle The informed set volume ratio decrease as $ \cbest $ decreases in the planning process in different dimensions.}
		\label{fig:sampling_efficiency:levelset}
	\end{subfigure}
	\begin{subfigure}[t!]{0.325\textwidth}
		\centering
		\includegraphics[width=\linewidth]{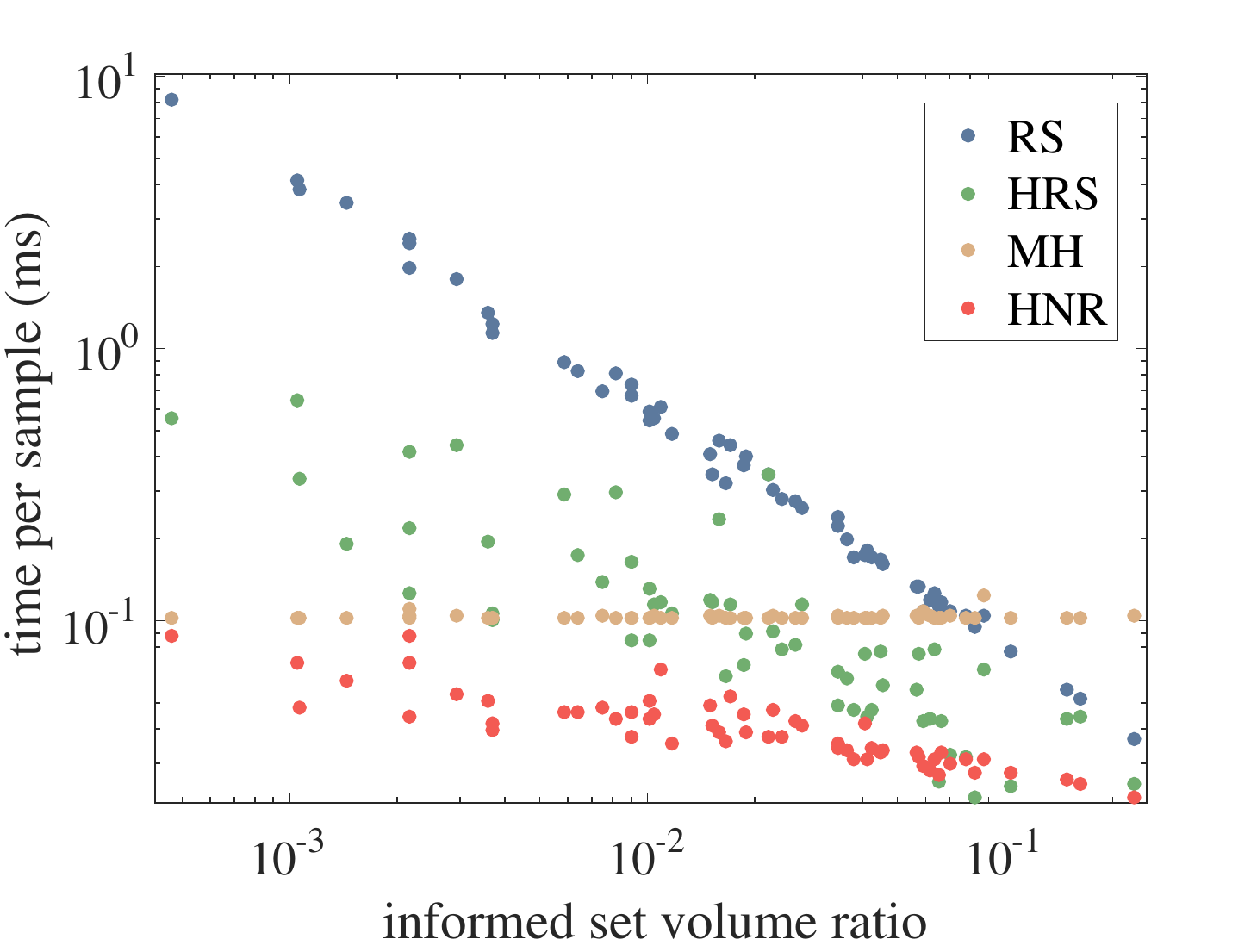}
		\caption{\captionstyle 4-dimension sampling space.}
		\label{fig:sampling_efficiency:2d}
	\end{subfigure}
	\begin{subfigure}[t!]{0.325\textwidth}
		\centering
		\includegraphics[width=\linewidth]{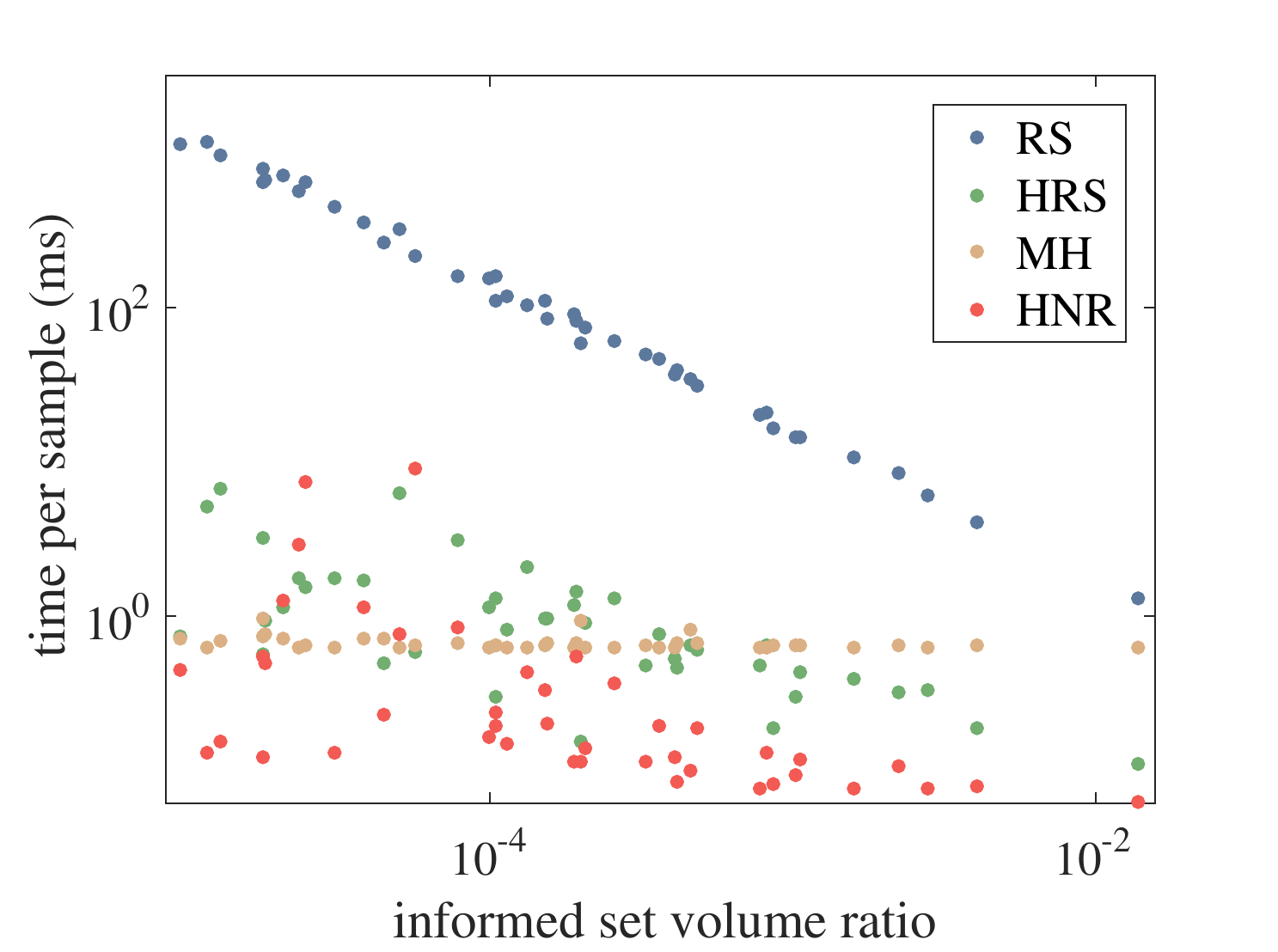}
		\caption{\captionstyle 12-dimension sampling space.}
		\label{fig:sampling_efficiency:6d}
	\end{subfigure}
	\caption{\captionstyle Average sampling time vs informed set volume ratio of four samplers (RS, HRS, MH and HNR) in state spaces of different dimensions. 
	The X axis is the ratio of informed set volume to the entire sampling space.
	The Y axis is the average time per sample.}
	\label{fig:sampling_efficiency}
\end{figure*}

Fig.~\ref{fig:sampling_efficiency:levelset} shows how the informed set volume ratio decreases as the informed set cost $ \cbest $ becomes smaller in problems of different dimensions.
In higher dimensions, the informed set volume ratio decreases much more quickly with decrease in the informed set cost $ \cbest $, as new cheaper trajectories are found in the planning process.

Fig.~\ref{fig:sampling_efficiency} shows the plot of the average time taken to generate one sample in the informed space vs. informed set volume ratio, i.e. the ratio of the volume of informed space to the volume of entire state space, for 5000 samples. 
The informed set volume ratio is estimated by the acceptance rate of rejection sampler.
The informed set volume ratio is approximated by the ratio of the number of accepted to the total number of samples obtained while running rejection sampling. 
Fig.~\ref{fig:sampling_efficiency} shows that MH and HNR have a better sampling efficiency compared to HRS and RS with decrease in informed set volume ratio or increase in dimensions. 

Metropolis-Hastings shows consistent sampling time when problems get harder. 
It takes the advantage of sampling a near state that generate samples in an informed set.
However this does not reflect the quality of the samples, though all the samples are in the informed set.
Recall in Algorithm~\ref{alg:mh_mcmc}, a new sample candidate is obtained from a Gaussian distribution $ q( x \mid x_{i-1}, \Sigma ) $.
The best covariance $ \Sigma $ that generates faster convergence differs with problem setting.
A small covariance tends to generate more samples near previous samples, while a large covariance has better exploration but is more likely to drive a Markov chain outside the informed set.
In our next planning experiment setting, we use the same covariance for different problems.

When informed set volume ratio is relatively high, it is easy to generate samples in the informed set.
All the samplers have close performances.
It actually implies rejection sampler is the best because of its simplicity in implementation and minimum correlation between successive samples.
The sampling time of all samplers except MH, increases as problems gets harder.
Notice that the sampling efficiency of HNR scales better than HRS and HRS is scales than RS.
Moving from a 4 dimension problem in Fig.~\ref{fig:sampling_efficiency:2d} to a 12 dimension problem in Fig.~\ref{fig:sampling_efficiency:6d}, sampling in an informed set becomes even harder, because the informed set volume ratio becomes smaller.
Here, HNR and MH samplers show much better efficiency over the others.

We want to point out that efficiently sampling in an informed set is not sufficient for determining the performance of a sampler.
For example, a sampler that constantly returns the same sample in a informed set might show the best sampling efficiency, however it is the worst sampler in a path planning problem.
Ideally, we want generated samples to be uniformly distributed in an informed set to get the best exploration.

\subsection{Planning Efficiency}

The quality of samples determines the efficiency of resulting planning algorithms.
If a sampler could provide samples with same quality as others but generate samples in a much efficiency way, we would expect that an informed RRT* with this sampler would show two properties.
\begin{itemize}
	\item It shall converge faster in finding the optimal solution.
	Sampling in an informed set is gradually becoming harder as new better solution reduces $ \cbest $ which reduces the informed set volume ratio.
	\item Its performance should not degrade significantly in high dimensional problems.
	As shown in Fig.~\ref{fig:sampling_efficiency:levelset}, the informed set volume ratio decrease more significantly in a high-dimension state space.
	The advantage of a good informed sampler becomes evident.
\end{itemize}

To evaluate the planning efficiency of the samplers, we run them with the informed RRT* planner~\cite{GSB14} in position-velocity space with MTDI as steering function, on three different problems described below and shown in Fig.~\ref{fig:problems}. 
For each problem the start and goal states (positions and velocities) are known in the joint space. 
Joint velocities at start and goal are calculated from desired end-effector velocities using inverse kinematics before starting the planning. Table~\ref{tab:params} shows the parameters used in the problems.

\begin {table*}[t!]
\centering
\setlength{\tabcolsep}{0.5em}
\begin{tabular}{r*{7}{>{[}r@{,~}l<{]}}} 
\toprule
HERB Joint & 
\multicolumn{2}{c}{1} & 
\multicolumn{2}{c}{2} & 
\multicolumn{2}{c}{3} &
\multicolumn{2}{c}{4} &
\multicolumn{2}{c}{5} &
\multicolumn{2}{c}{6} &
\multicolumn{2}{c}{7} 
\\ 
\midrule
Joint limits ($rad$)& 
 0.54 & 5.74 &
-2.00 & 2.00 &
-2.80 & 2.80 &
-0.90 & 3.10 &
-4.76 & 1.24 &
-1.60 & 1.60 &
-3.00 & 3.00
\\
$| v_{max} |~(rad/s)$ &
\multicolumn{2}{c}{0.75} &
\multicolumn{2}{c}{0.75} &
\multicolumn{2}{c}{2.00} &
\multicolumn{2}{c}{2.50} &
\multicolumn{2}{c}{2.50} &
\multicolumn{2}{c}{2.50} &
\multicolumn{2}{c}{2.00}
\\
\bottomrule
\end{tabular}
\begin{tabular}{>{[}r@{,~}l<{]}} 
\toprule
\multicolumn{2}{c}{3D Arm} 
\\ 
\midrule
 $-\pi$ & $\pi$ 
\\
\multicolumn{2}{c}{10} 
\\
\bottomrule
\end{tabular}
\begin{tabular}{>{[}r@{,~}l<{]}} 
\toprule
\multicolumn{2}{c}{6D Snake} 
\\ 
\midrule
 $-\pi$ & $\pi$ 
\\
\multicolumn{2}{c}{10} 
\\
\bottomrule
\end{tabular}
\caption {\captionstyle Parameters for the problems. All robots have $|a_{max}| = 1.0~rad/s^2$}.
\label{tab:params} 
\end{table*}

\subsubsection{Problem 1: 6 Dimension - 3 DoF Planar Manipulator}

The start and the goal states have zero velocities.
Fig.~\ref{fig:planning_efficiency:3dof:example} shows the planned path.

\subsubsection{Problem 2: 12 Dimension - 6 DoF Snake Arm}

The objective is to hammer the end-effector into the wall while starting with zero velocity.
Fig.~\ref{fig:planning_efficiency:6dof:example} shows the planned path.

\subsubsection{Problem 3: 14 Dimension - 7 DoF WAM Arm}

\begin{figure*}[t!]
	\centering
	\begin{subfigure}[b]{\textwidth}
	    \centering
		\includegraphics[height=2.45cm]{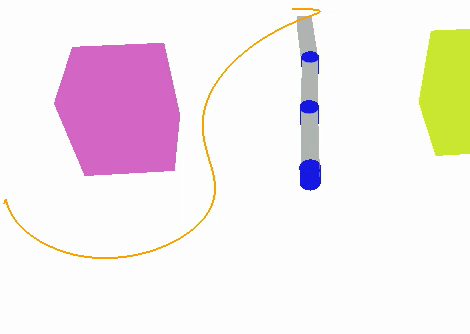}
		\includegraphics[height=2.45cm]{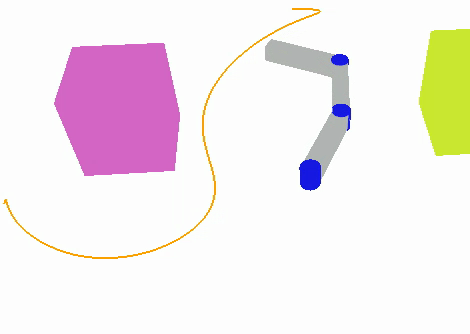}
		\includegraphics[height=2.45cm]{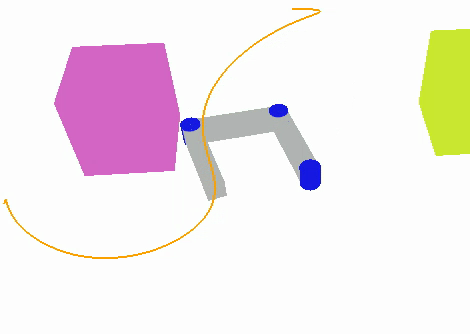}
		\includegraphics[height=2.45cm]{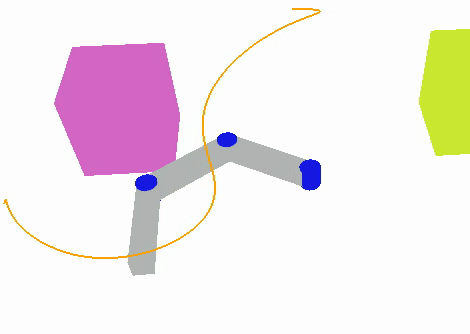}
		\includegraphics[height=2.45cm]{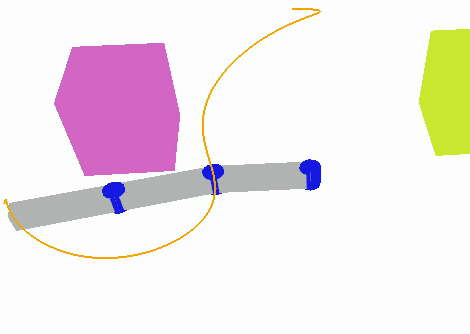}
		\caption{\captionstyle Problem 1 : 3DOF planar arm move from a start to a goal while starting and ending with zero velocities.}
		\vspace{6pt}
		\label{fig:planning_efficiency:3dof:example}
	\end{subfigure}
	\begin{subfigure}[b]{\textwidth}
	    \centering
		\includegraphics[height=3cm]{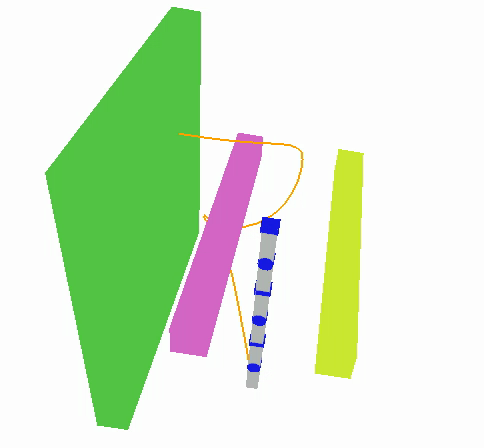}
		\includegraphics[height=3cm]{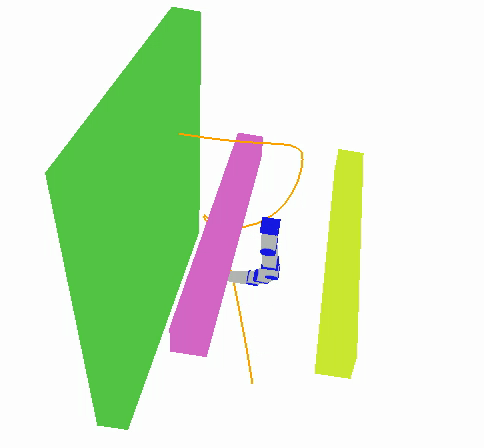}
		\includegraphics[height=3cm]{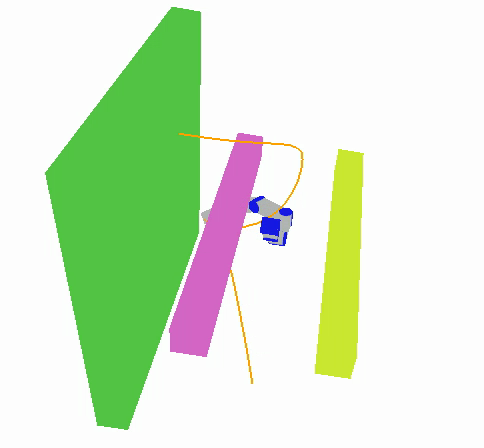}
		\includegraphics[height=3cm]{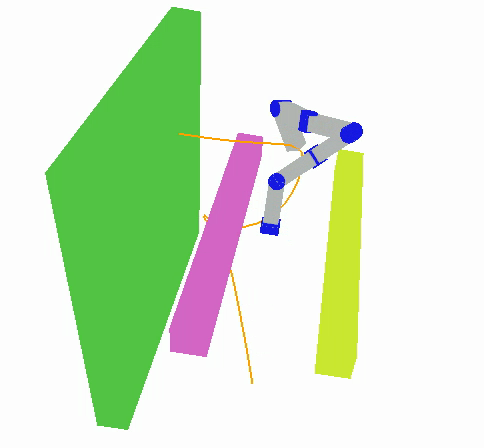}
		\includegraphics[height=3cm]{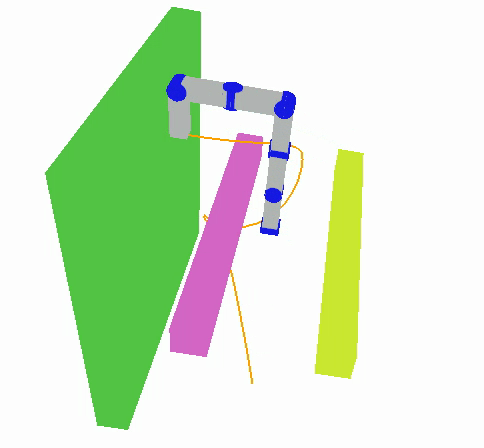}
		\caption{\captionstyle Problem 2 : 6DOF snake hammers the end-effector into the wall while starting with zero velocity.}
		\vspace{6pt}
		\label{fig:planning_efficiency:6dof:example}
	\end{subfigure}
	\begin{subfigure}[b]{\textwidth}
    \centering
    \includegraphics[height=2.7cm]{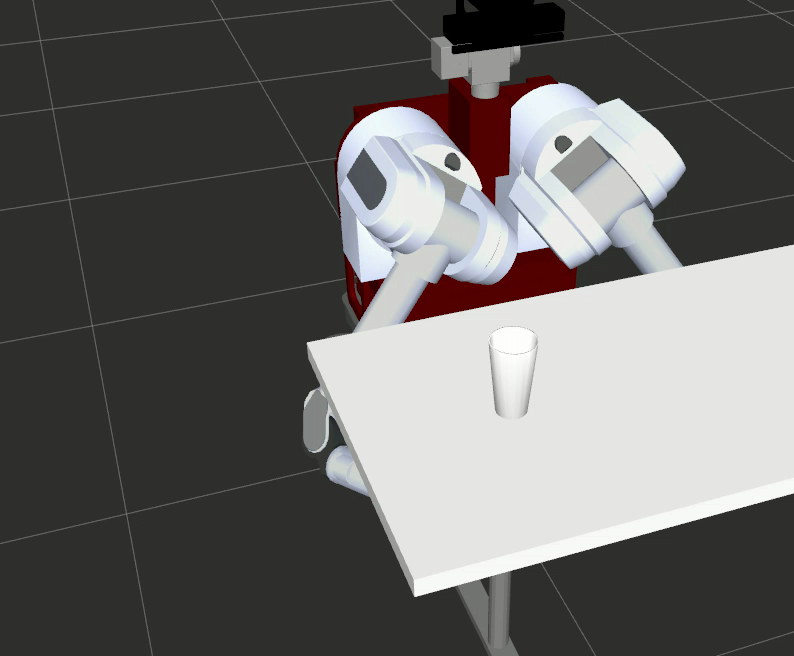}
    \includegraphics[height=2.7cm]{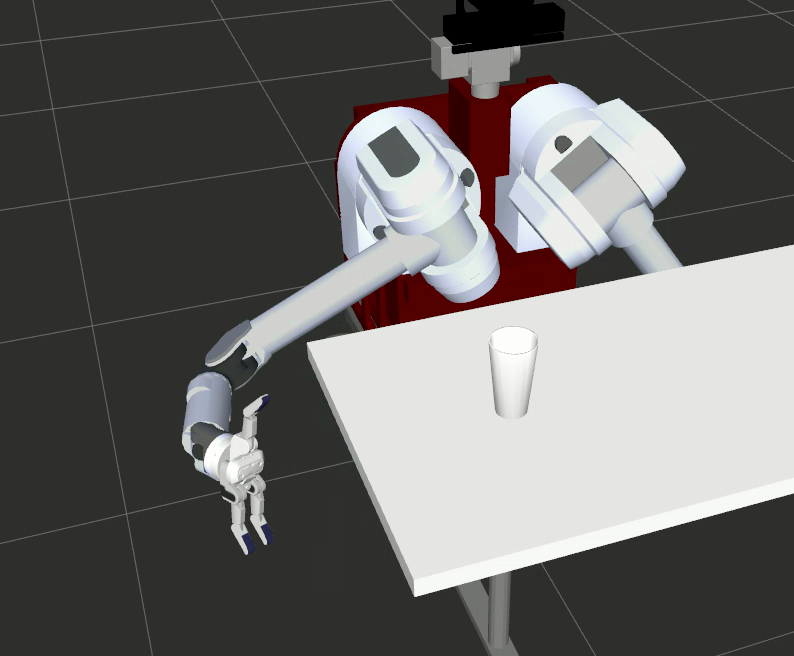}
    \includegraphics[height=2.7cm]{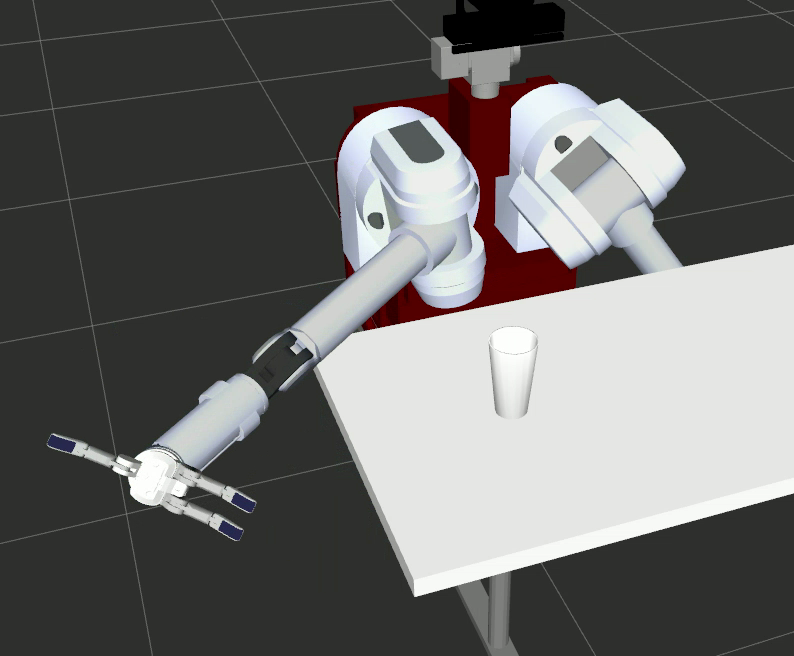}
    \includegraphics[height=2.7cm]{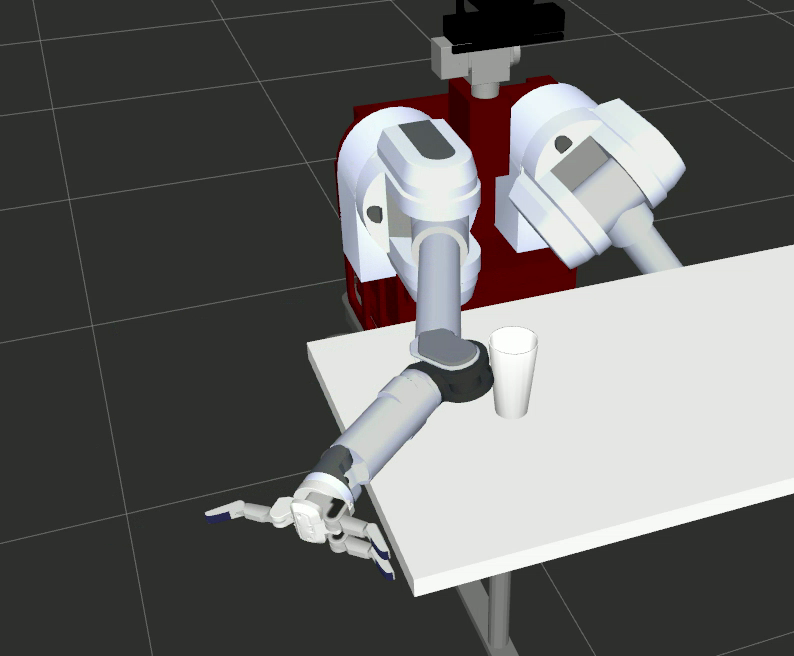}
    \includegraphics[height=2.7cm]{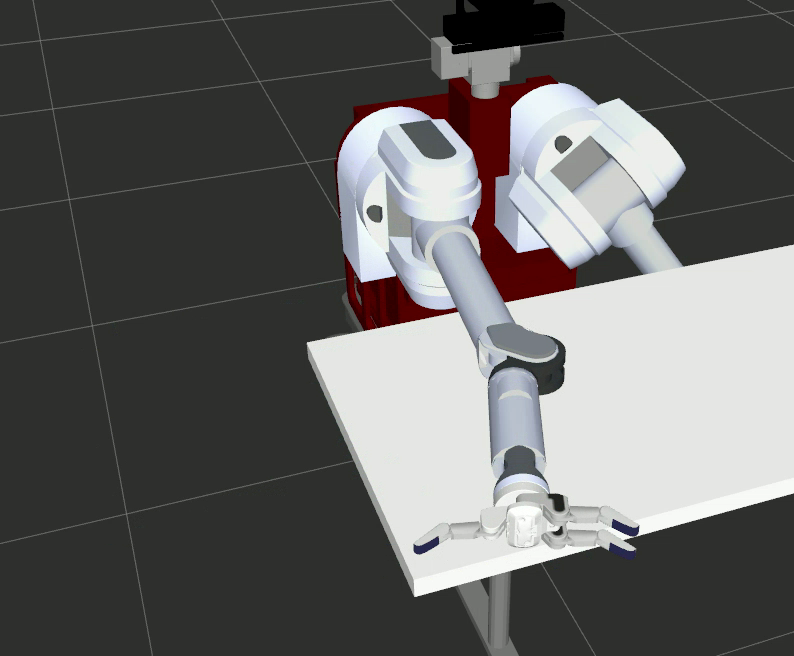}
	\caption{Problem 3 : HERB sweeps a cup on a table, in which the right arm starts with zero velocity and ends with non-zero velocity.}
	\label{fig:planning_efficiency:herb:example}
	\end{subfigure}
	\caption{\captionstyle Three problems are used to evaluate planning efficiency.
	These problems are defined in state spaces of different dimensions and subject to different kinodynamics.}
	\label{fig:problems}
\end{figure*}

The objective is to quickly swing away a glass on a table using the right arm.
Fig.~\ref{fig:planning_efficiency:herb:example} shows a few steps of a planned path.

\begin{figure*}[t!]
	\centering
	\begin{subfigure}[b]{0.325\textwidth}
		\includegraphics[width=\linewidth]{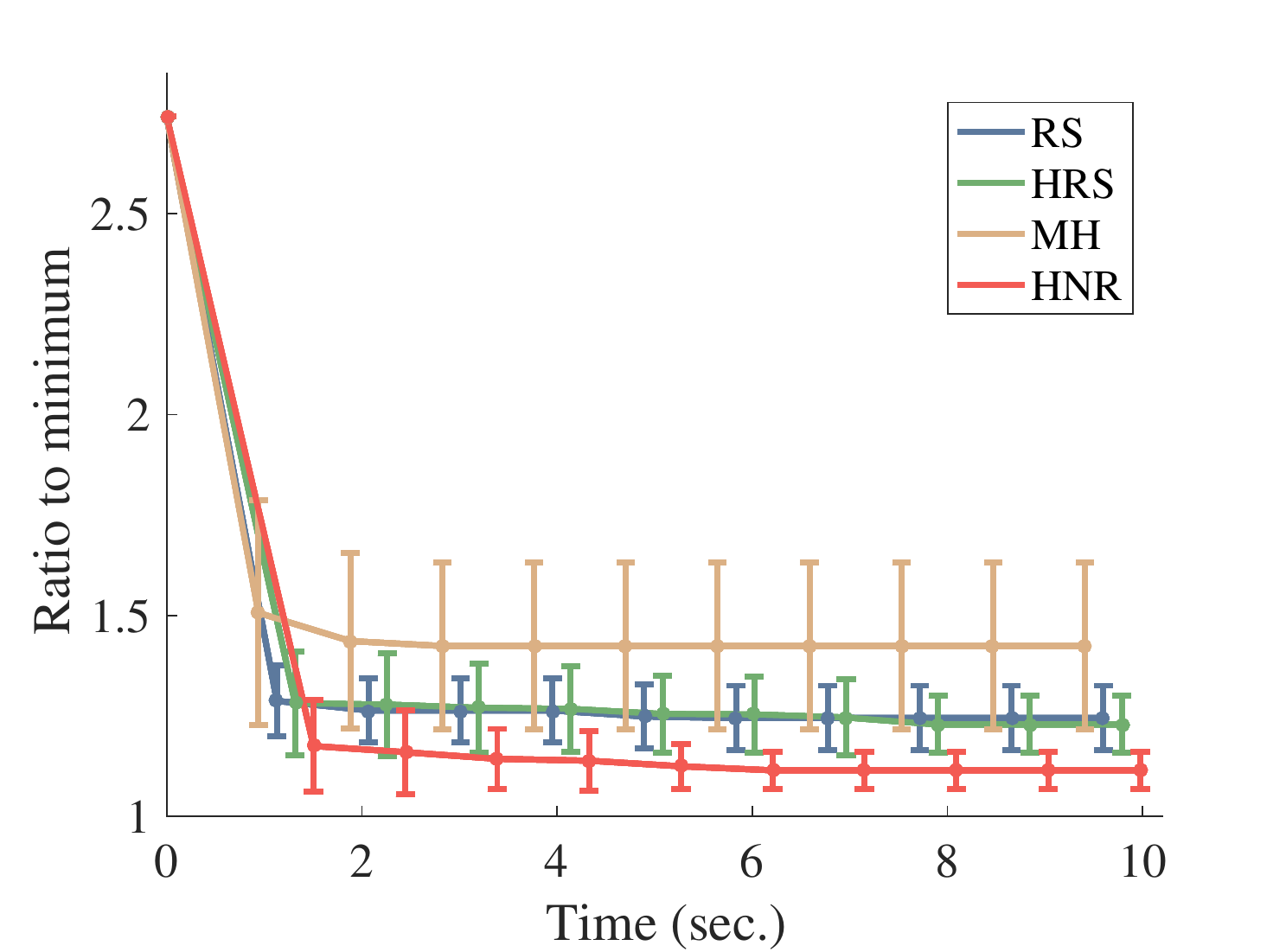}
		\caption{\captionstyle Problem 1 - 6 Dimensions.}
		\label{fig:planning_efficiency:3dof:general}
	\end{subfigure}	
	\begin{subfigure}[b]{0.325\textwidth}
		\includegraphics[width=\linewidth]{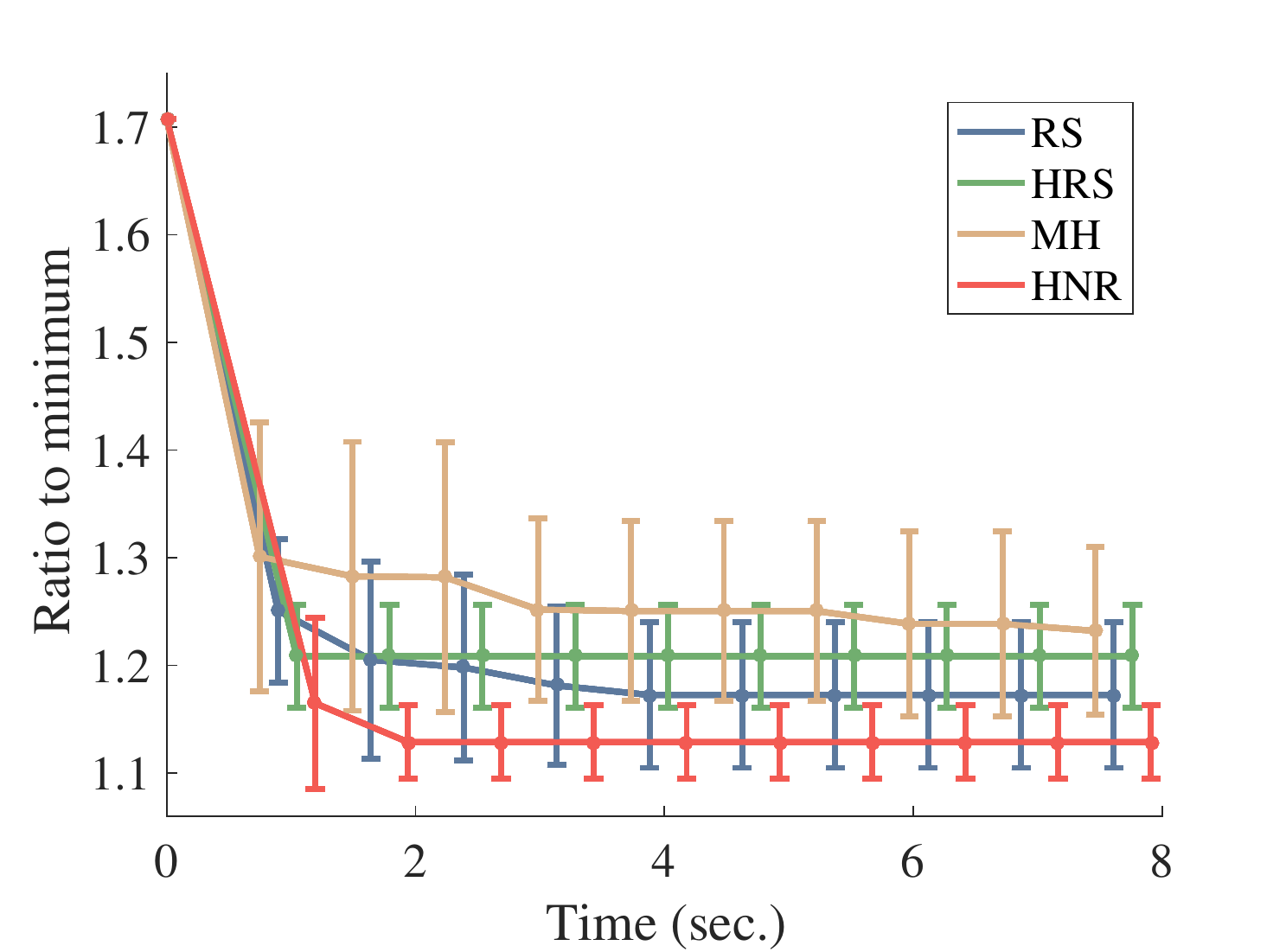}
		\caption{\captionstyle Problem 2 - 12 Dimensions.}
		\label{fig:planning_efficiency:6dof:hammering}
	\end{subfigure}
	\begin{subfigure}[b]{0.325\textwidth}
	\includegraphics[width=\linewidth]{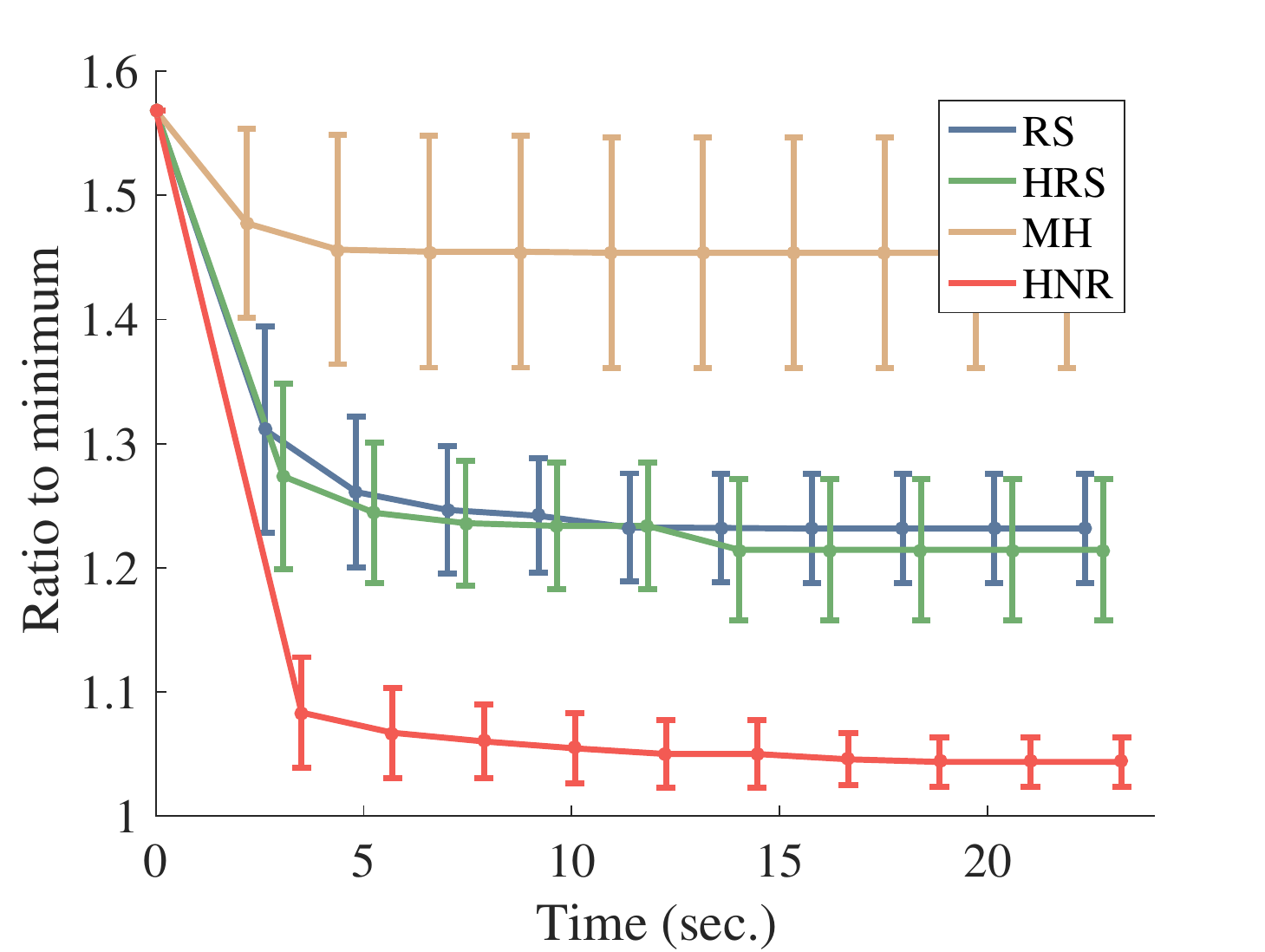}
	\caption{\captionstyle Problem 3 - 14 Dimensions.}
	\label{fig:planning_efficiency:herb:batting}
    \end{subfigure}
	\caption{\captionstyle Planning Efficiency of four different samplers (RS, HRS, MH and HNR) in three problems. 
	The X axis is the planning time. 
	The Y axis is the ratio of the current best and the optimal $ \cbest / c^*_{\rm best} $.}
	\label{fig:planning_efficiency}
\end{figure*} 


As shown in in Fig.~\ref{fig:planning_efficiency}, MH has the worst performance in all three problems, especially when the dimension increases.
Though theoretically samples converge to a target distribution only in the limit of infinite time. However, in practice the samples are to close to each other and don't explore the entire informed space.
If the variance of transition distribution is too high, it will tend to move out of the informed set too frequently, and takes longer to converge as the rejection rate is too high.

HNR shows close performance with RS and HRS in a 6-dimension problem, as in Fig.~\ref{fig:planning_efficiency:3dof:general}. 
As shown in Fig.~\ref{fig:planning_efficiency:6dof:hammering} and \ref{fig:planning_efficiency:herb:batting}, the advantages of HNR are clearly evident in higher dimensional problems. The cost of best solutions generated by planner with HNR sampler converges significantly faster to a cheaper to trajectory compared to others.

\section{Conclusion}
\label{sec:future}

In this work we demonstrated the effectiveness of using MCMC algorithms to 
efficiently produce samples for asymptotically-optimal motion planning algorithms.
Clearly, there are multiple other MCMC algorithms that can be used and it is interesting to see if alternative algorithms may produce better results.
One drawback of these approaches is that they usually incur parameters that have to be tuned. Indeed, in this work we did not spend effort in tuning the parameters and did not change them across the range of scenarios we tested. 
There is a wealth of literature in the optimization community regarding this topic and integrating  such tools is left for future work.
Finally, we are interested in using this framework with alternative sampling-based algorithms such as BIT*~\cite{GSB15} or LBT-RRT~\cite{SH16} and with alternative state spaces.

\section{Acknowledge}

This work was (partially) funded by the National Science Foundation IIS ($\#1409003$), and the Office of Naval Research.


\bibliography{bibliography}
\bibliographystyle{hieeetr}

\end{document}